\pdfoutput=1
\documentclass[10pt,twocolumn,letterpaper]{article}

\usepackage{iccv}              

\usepackage{graphicx}
\usepackage{afterpage}
\definecolor{iccvblue}{rgb}{0.21,0.49,0.74}
\usepackage[pagebackref,breaklinks,colorlinks,allcolors=iccvblue]{hyperref}
\usepackage{booktabs}
\usepackage{adjustbox}
\usepackage{multirow}
\usepackage{xcolor}
\usepackage{array}
\usepackage{wrapfig}
\usepackage{footmisc}

\def\paperID{9431} 
\def\confName{ICCV}
\def\confYear{2025}

\title{Flash Sculptor: Modular 3D Worlds from Objects}

\author{ Yujia Hu \quad Songhua Liu \quad Xingyi Yang \quad
Xinchao Wang \\ National University of Singapore\\
{ \tt \small \{yujia.hu, songhua.liu, xyang\}@u.nus.edu xinchao@nus.edu.sg }
}
\begin{document}

\twocolumn[{%
\renewcommand\twocolumn[1][]{#1}%
\maketitle
\vspace{-30pt}
\begin{center}
\centering
\includegraphics[width=1.0\linewidth]{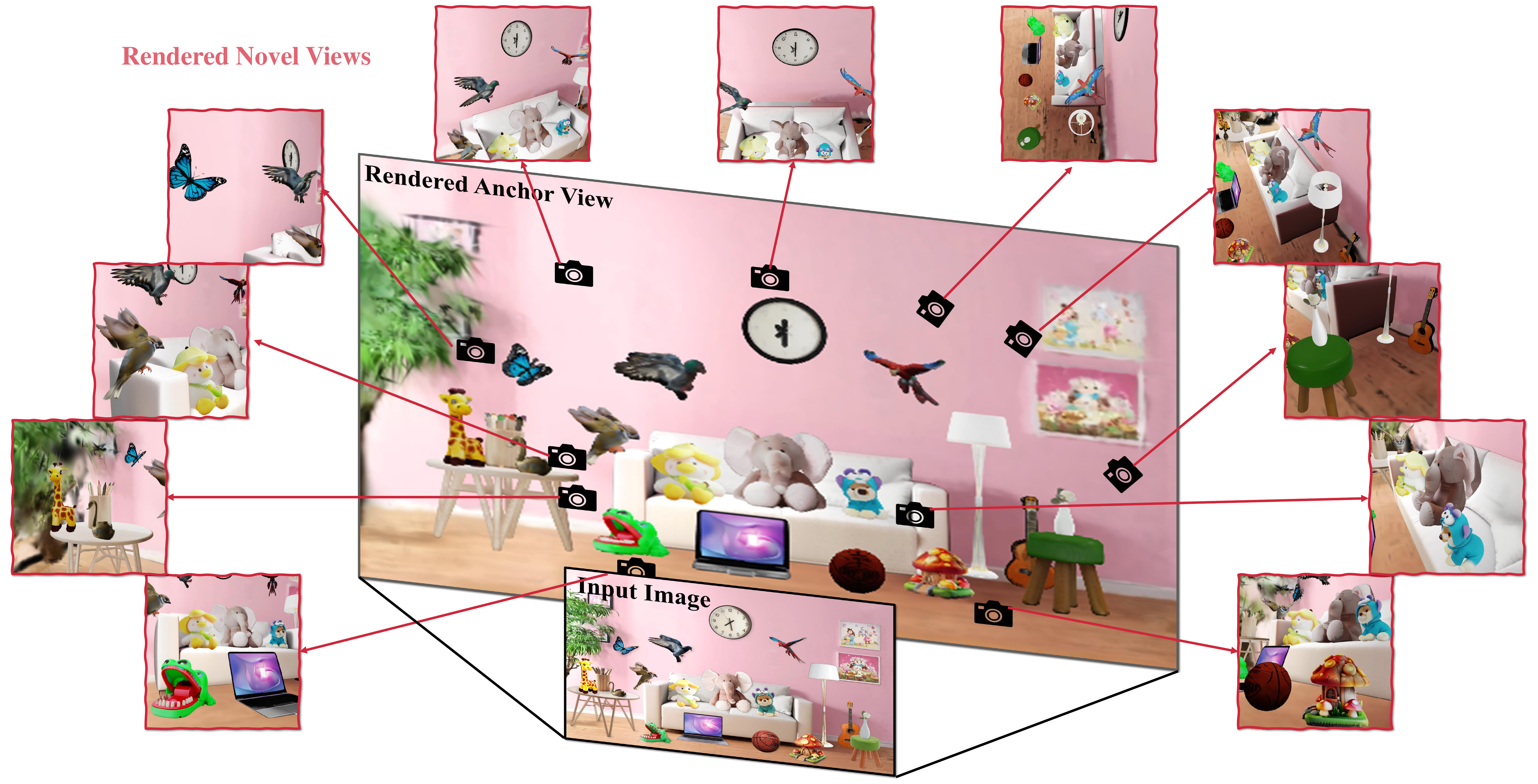}
\vspace{-15pt}
\captionof{figure}{A scene generated by our method. Our method can generate high-quality 3D scene from a single image. 
}
\label{fig: teaser}
\end{center}%
}]


\begin{abstract}
Existing text-to-3D and image-to-3D models often struggle with complex scenes involving multiple objects and intricate interactions. 
Although some recent attempts have explored such compositional scenarios, they still require an extensive process of optimizing the entire layout, which is highly cumbersome if not infeasible at all. 
To overcome these challenges, we propose Flash Sculptor in this paper, a simple yet effective framework for compositional 3D scene/object reconstruction from a single image. 
At the heart of Flash Sculptor lies a divide-and-conquer strategy, which decouples compositional scene reconstruction into a sequence of sub-tasks, including handling the appearance, rotation, scale, and translation of each individual instance. 
Specifically, for rotation, we introduce a coarse-to-fine scheme that brings the best of both worlds—efficiency and accuracy—while for translation, we develop an outlier-removal-based algorithm that ensures robust and precise parameters in a single step, without any iterative optimization. 
Extensive experiments demonstrate that Flash Sculptor achieves at least a $3\times$ speedup over existing compositional 3D methods, while setting new benchmarks in compositional 3D reconstruction performance. Codes are available \href{https://github.com/YujiaHu1109/Flash-Sculptor}{here}.
\end{abstract}    
\section{Introduction}
\label{sec:intro}

Reconstructing 3D scenes from a single image or prompt remains a fundamental challenge in computer vision, particularly for scenes that contain multiple objects with complex interactions. Although recent text-to-3D \cite{chen2024text,jiang2024general} and image-to-3D \cite{tang2024lgm,xiang2024structured} models have leveraged advanced diffusion techniques \cite{rombach2022high} and 3D Gaussian Splatting \cite{kerbl20233d} to achieve promising results in single-object reconstruction, these methods often falter in compositional settings. In particular, accurately reconstructing entangled instances and precisely capturing the spatial relationships among objects remain significant hurdles, as the intricate interplay between objects can easily lead to erroneous or incoherent reconstructions.

To address these issues, compositional 3D generation has emerged as a promising solution for synthesizing multi-object scenes. However, existing approaches typically rely on iterative optimization techniques to determine the overall layout of various instances. 
For example, some methods leverage Score Distillation Sampling (SDS) \cite{chen2024comboverse,poole2022dreamfusion} or differentiable layout alignment \cite{han2024reparo} to optimize object placement and spatial configuration, while others \cite{ge2024compgs,zhou2024gala3d} dynamically refine 3D Gaussian representations initialized from 2D compositional priors. Furthermore, several approaches \cite{cheng2023progressive3d,zhou2024layout} adopt multi-stage editing strategies that iteratively refine the layout of the whole scene. These holistic optimization paradigms are not only computationally intensive and cumbersome but also become practically infeasible when dealing with scenes containing a large number of interacting objects, especially for end users. 

To overcome these challenge, in this paper, we propose \textit{Flash Sculptor}, a simple yet effective pipeline for compositional 3D scene reconstruction from a single image. 
Unlike previous approaches that optimize the entire layout, at the heart of Flash Sculptor lies in a divide-and-conquer strategy breaks the reconstruction task into a sequence of manageable sub-tasks—handling the appearance, rotation, scale, and translation of each individual instance. 
In this way, we bypass the complexity of determining object relationships as a whole and enhance the efficiency significantly. 
Moreover, the processing of individual instances can potentially be parallelized, further enhancing practical scalability. 

Since there is no explicit operation addressing the holistic relationships in a complex scene, it highlights the importance of carefully handling the appearance and spatial parameters of each individual component. 
To this end, we back up the proposed Flash Sculptor with two innovative strategies, each designed to estimate rotation and translation parameters, respectively. 
On the one hand, for the rotation estimation stage, we introduce a coarse-to-fine scheme that first provides a rapid, approximate estimation of the rotation parameters, and then adopts a simplex method \cite{singer2009nelder} to refine the rotation. On the other hand, for translation, we develop an outlier-removal algorithm that robustly computes the depth-related shift by mitigating the impact of erroneous depth estimates or mismatches between 2D and 3D correspondences, resulting in accurate translation parameters in a single optimization step. 

We validate the effectiveness of Flash Sculptor via extensive experiments on both image and text benchmarks. Our results show significant improvements in accuracy and time efficiency compared to existing image-to-3D and text-to-3D models, as well as recent compositional 3D generation methods. Specifically, our method achieves a $3\times$ speedup over state of the arts, while setting new records on {T$^{3}$Bench}. 
Our main contributions are summarized as follows:
\begin{itemize}
    \item We propose \textit{Flash Sculptor}, a simple yet effective pipeline for compositional 3D scene reconstruction. To the best of our knowledge, we are the first to incorporate scene reconstruction to compositional 3D generation from a single image. 
\item By leveraging the divide-and-conquer strategy, we decouple the estimation of appearance, rotation, scale, and translation for each object and devise effective strategies for each stage, thereby reducing the need for exhaustive global layout optimization. 
    \item Extensive experiments demonstrate that our method can generate state-of-the-art results of compositional scenes from a single image with $3\times$ acceleration compared with previous methods and can be readily applied to downstream tasks like 3D editing.
\end{itemize}

\section{Related Work}

\begin{figure*}[ht]
    \centering
    \includegraphics[width=1.0\linewidth]{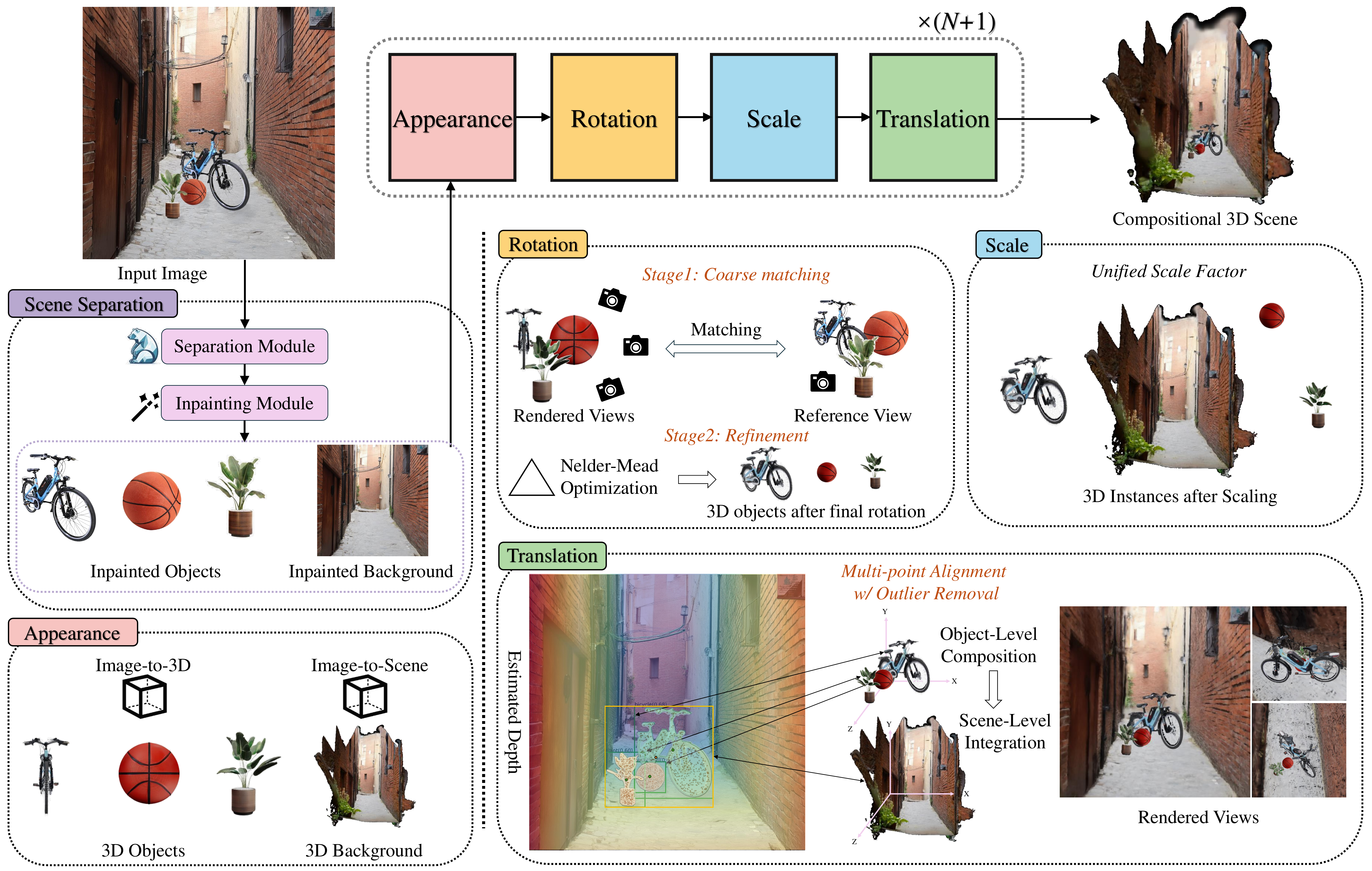}
    \vspace{-.2in}
    \caption{Overview of our method. Given an input image, our method first separate the image into \(N\) independent objects and a background. Then, all these instances undergoes disentangled four stages, i.e., appearance, rotation, scale and translation. Finally, after finishing these sub tasks, our pipeline generates a compositional 3D scene consistent with the original input.}
    \label{fig:pipeline}
    \vspace{-.1in}
\end{figure*}

\noindent \textbf{Text-to-3D and Image-to-3D Generation.} Recent advances in 3D generation have been largely driven by the advent of powerful diffusion models \cite{chen2024v3d,rombach2022high}. 
In text-to-3D domain, DreamFusion \cite{poole2022dreamfusion} introduced Score Distillation Sampling (SDS) to harness the capability of 2D diffusion models for 3D synthesis. 
Subsequent studies \cite{wang2023prolificdreamer,yu2023text} refined SDS methods and leveraged advanced 3D representations such as 3D Gaussian Splatting (3DGS) \cite{kerbl20233d} to enhance both quality and efficiency. 
In parallel, image-to-3D reconstruction methods focus on ensuring geometric and textural consistency between generated assets and input images. Early approaches \cite{melas2023realfusion,tang2023make,liu2023zero} employed 2D diffusion models \cite{rombach2022high} as priors for 3D representation learning, while more recent paradigms, such as those based on fine-tuned diffusion models \cite{yang2024hi3d} and feed-forward architectures \cite{hong2023lrm,tang2024lgm}, have significantly improved the multi-view consistency of image-to-3D models. 
Despite these impressive strides, most of current text-to-3D and image-to-3D models are trained on single-object datasets \cite{deitke2023objaverse,yu2023mvimgnet}, thereby challenges remain in generating compositional 3D scenes that are both coherent and plausible.

\noindent \textbf{Compositional 3D Generation.} 
Recent work in compositional 3D generation has sought to address the challenges of synthesizing complex scenes by jointly modeling individual 3D assets and their spatial arrangements. However, a critical limitation of existing methods is their reliance on extensive optimization of the entire scene layout. For example, in image-to-3D domain, ComboVerse \cite{chen2024comboverse} employs spatially-aware diffusion guidance, whereas REPARO \cite{han2024reparo} leverages differentiable 3D layout alignment. Similarly, in text-to-3D realm, approaches such as CompGS \cite{ge2024compgs} and GALA3D \cite{zhou2024gala3d} dynamically optimize 3D Gaussian representations initialized from 2D compositional priors, while others \cite{po2024compositional,zhou2025layoutdreamer} depend on optimizing comprehensive spatial configurations guided by physics or semantic cues. Certain methods \cite{zhou2024layout,cheng2023progressive3d} further exacerbate this issue by adopting multi-stage editing strategies that refine full layouts. These global optimization paradigms are highly cumbersome and even become practically infeasible as the number of interacting objects increases. In this paper, we disentangle the process of optimizing the entire scene layout by independently solving for the appearance, rotation, scale, and translation of each instance, thereby achieving significantly higher efficiency while maintaining satisfactory performance.


\noindent \textbf{3D Scene Generation.} Generating 3D scenes from a single image is a long-standing challenge in computer vision and graphics. NeRF-based methods \cite{mildenhall2021nerf,muller2022instant,sitzmann2019scene} enable scene-level view synthesis but struggle with narrow baselines and limited parallax. 3D Gaussian Splatting (3DGS) approaches \cite{zhang2024towards} achieve high-fidelity reconstructions but are often object-centric.  
To enhance scene continuity and expand views, some methods utilize inpainting models \cite{shriram2024realmdreamer,chung2023luciddreamer} or video generation models \cite{you2024nvs,wang2024motionctrl} to iteratively complete missing regions. 
Some \cite{wang2024vistadream} further enhances 3D scene generation by fusing Vision-Language Models with multi-view sampling to enforce geometric coherence.
In our pipeline, unlike other compositional methods that focus solely on assembling object-level assets, we also leverage a scene generation model to synthesize the 3D background, which will be used later to combine with objects to achieve true compositional scene.

\section{Method}

In this section, we will introduce our divide-and-conquer pipeline in detail. In Sec.~\ref{sec:3.1}, we first separate the scene into objects and background. Then we come to the four sub-tasks. Sec.~\ref{sec:3.2} shows our strategies for reconstructing 3D appearance, Sec.~\ref{sec:3.3},  Sec.~\ref{sec:3.4}, Sec.~\ref{sec:3.5} details our strategies for determining the parameters for 3D combination, i.e., rotation \(r_i\), scale \(s_i\) and translation \(t_i\) respectively. The overall pipeline is illustrated in Fig.~\ref{fig:pipeline}.

\subsection{Scene Separation} \label{sec:3.1}

\textbf{Object Extraction.} Owing to the powerful text-to-image generation capabilities of models such as Stable Diffusion~\cite{von-platen-etal-2022-diffusers}, our approach can initiate from either text or image. Given a reference image $I_{ref}$, we employ Grounded SAM~\cite{ren2024grounded} to automatically detect and segment the objects, yielding the bounding boxes, masks and labels for each instance:
\begin{equation}
B_i, M_i, L_i = \text{Grounded SAM}(I_{\text{ref}}), \quad i \in \{1, 2, \dots, N\},
\end{equation}
where \(B_i\), \(M_i\), \(L_i\) represent the bounding box, mask and label of the \(i\)-th object in \(I_{\text{ref}}\) respectively, and there are \(N\) objects in total in \(I_{\text{ref}}\).

\begin{figure}[t]
    \centering
    \includegraphics[width=1.0\linewidth]{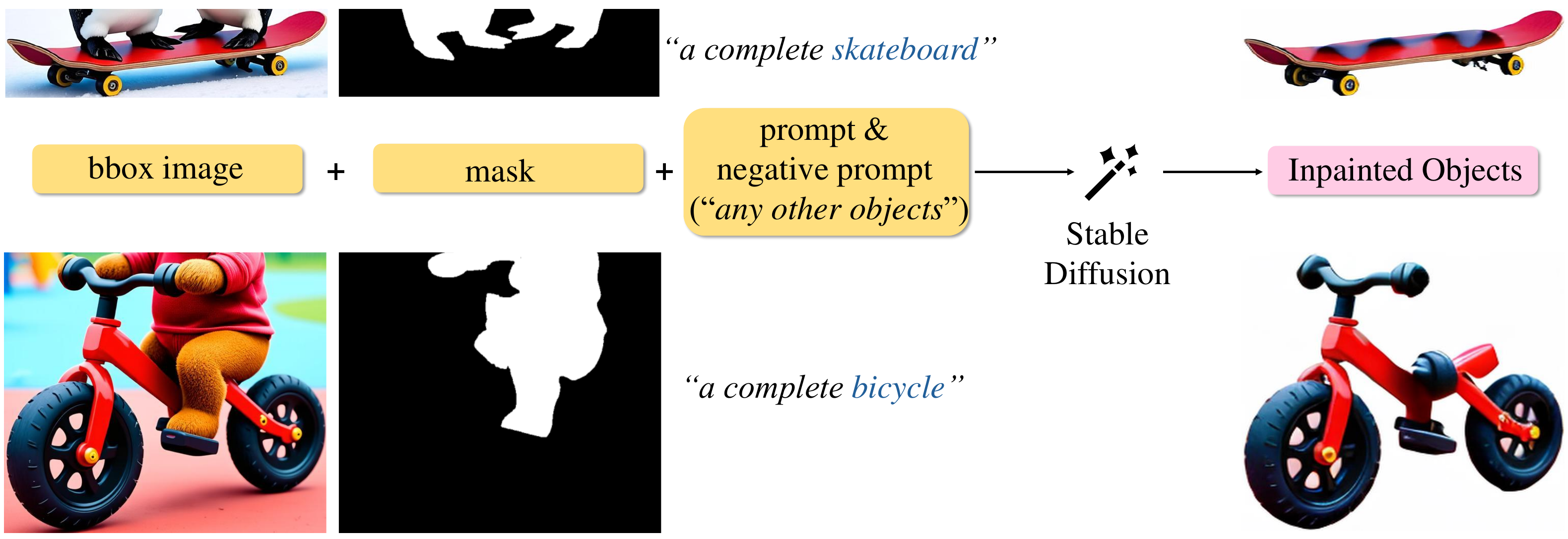}
    \vspace{-.2in}
    \caption{The object inpainting strategy of our method. We feed the bounding box image, mask and prompts to stable diffusion to generate inpainted objects.}
    \label{fig:mask}
    \vspace{-.15in}
\end{figure}

\begin{figure}[t]
    \centering
    \includegraphics[width=1.0\linewidth]{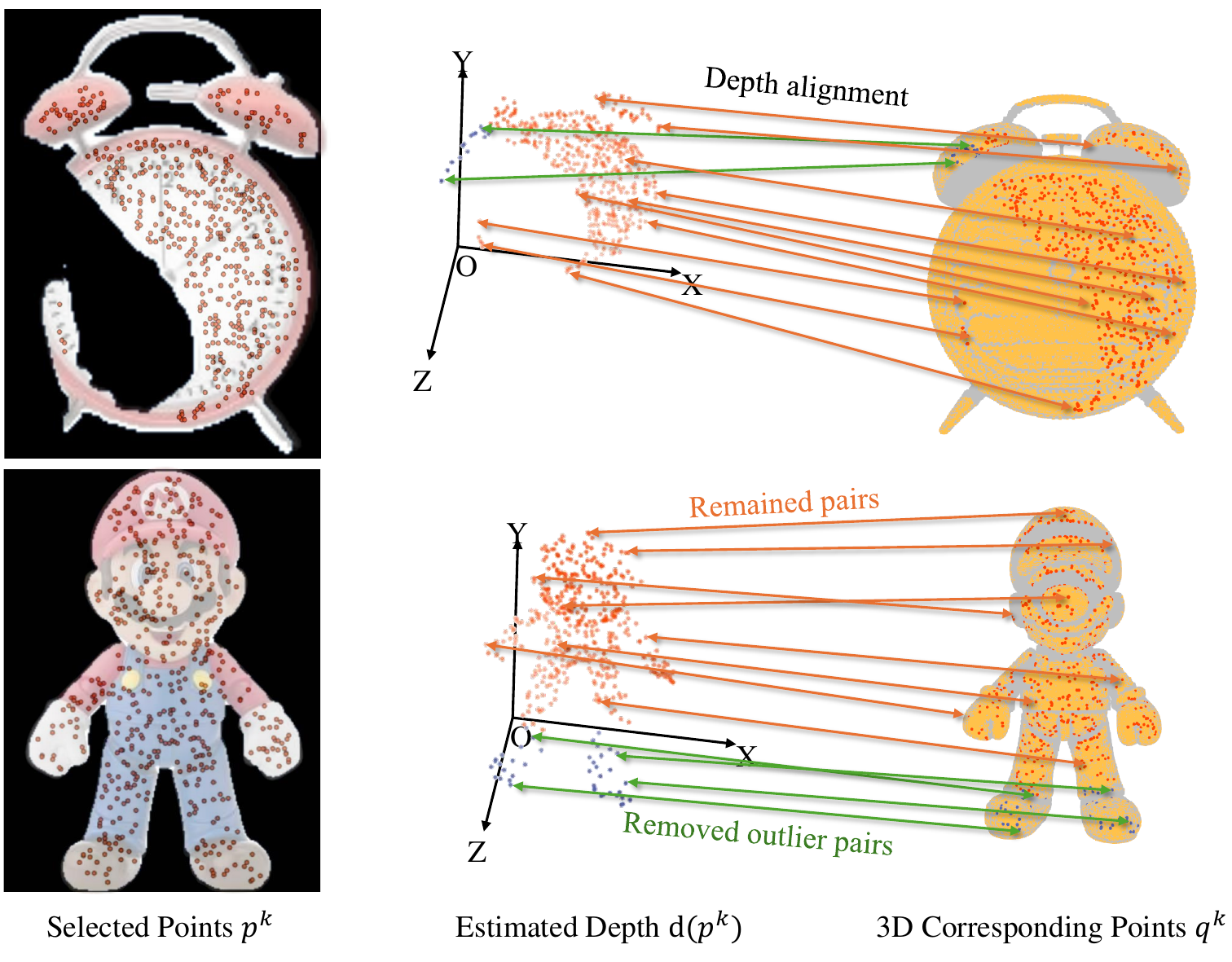}
    \vspace{-.2in}
    \caption{Illustration of the process of selecting points and removing outliers for depth alignment.}
    \label{fig:points}
    \vspace{-.15in}
\end{figure}

Since objects in composite scenes often occlude one another, an inpainting process is required to recover the occluded parts. For each object, we extract its bounding box \(B_i\) and provide the corresponding region to Stable Diffusion. However, as ground-truth masks for the original objects are not available, we design a mask computation strategy to maximize the integrity of each instance while excluding the target object’s own mask. Specifically, we define the inpainting mask for the \(i\)-th object as:
\begin{equation}
M_{inpaint_i} = \bigcup_{\substack{j \in \{1, 2, \dots, N\} \\ j \neq i,\, B_j \cap B_i \neq \emptyset}} M_j,
\end{equation}
where \(M_{inpaint_i}\) is the mask used during inpainting for the \(i\)-th object’s bounding box \(B_i\). This formulation ensures that all the relevant regions of other objects overlapping with \(B_i\) are included in the mask, as illustrated in Fig.~\ref{fig:mask}. Furthermore, to generate more ideal results, we provide Stable Diffusion with a positive text prompt \textquotedblleft\textit{a complete \(L_i\)}\textquotedblright\ and a negative prompt \textquotedblleft\textit{any other objects}\textquotedblright\ to better guide the synthesis of complete objects.

\noindent \textbf{Background Extraction.} To obtain a clean background \(I_{\text{bg}}\), we combine the object masks:
\begin{equation}    
M_{\text{objects}} = \bigcup_{i=1}^{N} M_i,
\end{equation}
and feed this composite mask into a mask-inpainting model~\cite{suvorov2022resolution} to inpaint the object regions, which results in a background image \( I_{\text{bg}} \) containing only the scene backdrop.

\subsection{Appearance} \label{sec:3.2}
\textbf{Single Object 3D Reconstruction.} With complete 2D object images obtained via inpainting, we reconstruct each object using current image-to-3D methods based on 3D Gaussian Splatting (3DGS)~\cite{tang2024lgm,xiang2024structured,yang2024hunyuan3d} to ensure seamless integration and high-quality rendering.

\noindent \textbf {3D Scene Reconstruction.} For background, we adopt VistaDream~\cite{wang2024vistadream} to generate a full 3D scene from \(I_{\text{bg}}\) as it can sample multiview consistent images for single-view scene reconstruction and guarantee that the initial view remains consistent with \(I_{\text{ref}}\).

\subsection{View-Consistent Rotation} \label{sec:3.3}

To obtain the rotation parameters \( r_{i} \) in an efficient way, our method begins by deriving a coarse initial estimation of the object’s orientation through feature alignment. Specifically, we render multiple views of a coarse 3D instance \( I_i \) under \(C^2\) distinct camera configurations, where the azimuth and elevation angles are uniformly sampled to form the set:
\begin{equation}
\Pi = \{(e_j, a_j) \mid j = 1, \dots, C\} \subset \mathbb{R}^{C \times 2}.
\end{equation}
For each object, we compute feature representations using DINOv2~\cite{oquab2023dinov2}, to extract the feature vector \( \mathbf{f}_j \) for each view and the reference patch \( \mathbf{f}_{ref} \) from the 2D instance \( \hat{I}_i \). The initial rotation parameters \( (e^*, a^*) \) are then determined by selecting the azimuth and elevation pair from the set \( \Pi \) that maximizes the cosine similarity between the reference patch \( \mathbf{f}_{\text{ref}} \) and the corresponding feature \( \mathbf{f}_j \) for each rendered view:
\begin{equation}
(e^*, a^*) = \arg\max_{(e,a) \in \Pi} \frac{1}{C^2}\sum_{j=1}^{C^2} \cos\left( \mathbf{f}_{\text{ref}}, \mathbf{f}_j \right),
\end{equation}
where \(C^2\) is the number of views. To refine the initial estimation, we further optimize the rotation parameters using the Nelder-Mead optimization~\cite{singer2009nelder} algorithm, which minimizes the SSIM~\cite{wang2004image} between the rendered views and $I_{ref}$, leading to a more precise alignment of the object’s orientation with its 2D counterpart.

\subsection{Unified Scale Calibration} \label{sec:3.4}
As for scale factor of each instance, unlike other methods that overlook the influence of depth information, we propose a unified scale calibration strategy that accounts for 2D size, 3D scale, and depth compensation to align the scale of 3D objects with their 2D counterparts. The unified scale factor for an object $\mathcal{O}_i$ is defined as:

\begin{equation}
s_i = \underbrace{\frac{\max(W_{B_i}, H_{B_i})}{\max\limits_j(\max(W_{B_j}, H_{B_j}))}}_{\text{2D normalization}} 
\times 
\underbrace{\frac{\|\mathcal{O}_i\|_{\text{diag}}} {\min\limits_j\|\mathcal{O}_j\|_{\text{diag}}}}_{\text{3D normalization}} 
\times 
\underbrace{\frac{D_i}{D_{\min}}}_{\text{depth compensation}},
\end{equation}

where $W_{B_i}$ and $H_{B_i}$ are the width and height of the bounding box $B_i$ of the $i$-th object, $\max(W_{B_j}, H_{B_j})$ is the maximum dimension among all object bounding boxes. $\|\cdot\|_{\text{diag}}$ denotes the diagonal length of the 3D mesh of the object $\mathcal{O}_i$, calculated as $\|\mathcal{O}_i\|_{\text{diag}} = \sqrt{W_{M_i}^2 + H_{M_i}^2}$, where $W_{M_i}$ and $H_{M_i}$ are the width and height of the $i$-th object’s 3D mesh. $D_i$ is the mean depth of the $i$-th object from the depth estimation model \cite{depth_anything_v2}, and $D_{\min}$ represents the minimum mean depth across all objects.

\subsection{Robust Translation Alignment} \label{sec:3.5}
To accurately align each 3D instance with its corresponding 2D region, we first compute raw translation estimates by establishing correspondences between the 2D image and the 3D model. The translations along the \(x\) and \(y\) axes are derived by aligning the centers of the 2D bounding box and the 3D instance. However, estimating the \(z\)-axis translation is more challenging due to the inherent ambiguity of depth in a single 2D image.

To obtain reliable depth translations, as illustrated in Fig. \ref{fig:points}, we sample \(K\) points \(\{p_k \in M_i\}_{k=1}^{K}\) from the object mask \(M_i\) and compute their estimated depths \(d(p_k)\) using a depth estimation model \cite{depth_anything_v2}. For each sampled point, we determine its nearest 3D surface correspondence \(q_k\) by first applying the HPR algorithm to extract the visible surface points of the object, and then identifying the 3D point that best matches the relative position of \(p_k\) within the bounding box. 

Owing the discrepancies between the 2D image space and the 3D coordinate system, we introduce axis-specific scaling factors \(\lambda_x\), \(\lambda_y\), and \(\lambda_z\). These factors are derived from a 3D scaffold of the reference image \(I_{\text{ref}}\). Rather than constructing the scaffold by computing a depth map and back-projecting pixels from scratch, we leverage our background scene reconstruction model~\cite{wang2024vistadream} to generate a complete 3D representation. From the resulting visible point cloud in the reference view, we compute the spatial extents along the \(x\), \(y\), and \(z\) axes, denoted as \(S_x^{\text{scaffold}}\), \(S_y^{\text{scaffold}}\), and \(S_z^{\text{scaffold}}\), respectively. These extents define the scaling factors:
\begin{equation}
\lambda_x : \lambda_y : \lambda_z = S_x^{\text{scaffold}} : S_y^{\text{scaffold}} : S_z^{\text{scaffold}}.
\end{equation}

Let \(z(q_k)\) denote the depth of \(q_k\). Then we can calculate the depth difference \(\Delta z_k\) for each point can be defined as the difference between \(d(p_k)\) and  \(z(q_k)\) after scaling.

Due to potential mismatches and erroneous depth estimates, outliers inevitably arise during alignment. To robustly mitigate their impact, we employ the Median Absolute Deviation (MAD) to filter out points with excessive deviation and obtain the final $\hat{K}$ pairs for alignment. We compute the MAD as:
\begin{equation}
\mathrm{MAD} = \mathrm{median}\Bigl(|\Delta z_k - \mathrm{median}(\Delta z_k)|\Bigr),
\end{equation}
and discard points satisfying \(|\Delta z_k - \mathrm{median}(\Delta z_k)| > \lambda\,\mathrm{MAD}\), where \(\lambda\) is a predetermined threshold. 

Finally, the scaled translation parameters for object \(\mathcal{O}_i\) are defined as:
\begin{equation}
\mathbf{t}_i =
\begin{cases}
t_x = \lambda_x\, c_x^{2D} - s_i\, c_x^{3D}, \\[1ex]
t_y = \lambda_y\, c_y^{2D} - s_i\, c_y^{3D}, \\[1ex]
t_z = \operatorname{median}\!\Bigl\{ \lambda_z\, d(p_k) - s_i\, z(p_k) : k = 1,\dots,\hat{K} \Bigr\}.
\end{cases}
\end{equation}

\noindent \textbf{Final Scene Composition.} When generating the whole scene using reconstructed 3D models, object-level composition and scene-level integration are unified into a single pipeline that jointly optimizes the rotation, scale, and translation of each 3D instance in our framework. Given a reference image \(I_{\text{ref}}\), we first align each individual 3D object with its corresponding 2D counterpart, thereby computing object-specific parameters—namely, the scale \(s_i\), rotation \(r_i\), and translation \(t_i\) , as described in Sections~\ref{sec:3.3}, \ref{sec:3.4}, \ref{sec:3.5}. Subsequently, the collection of aligned objects is treated as a single entity, and their global parameters are determined relative to the scene to ensure consistent integration of all instances. Notably, since the reference view has been pre-aligned during the scene integration stage, no further rotational adjustment is necessary when merging the composed objects into the global scene.

\section{Experiments}

\begin{figure*}[ht]
    \centering
    \includegraphics[width=1.0\linewidth]{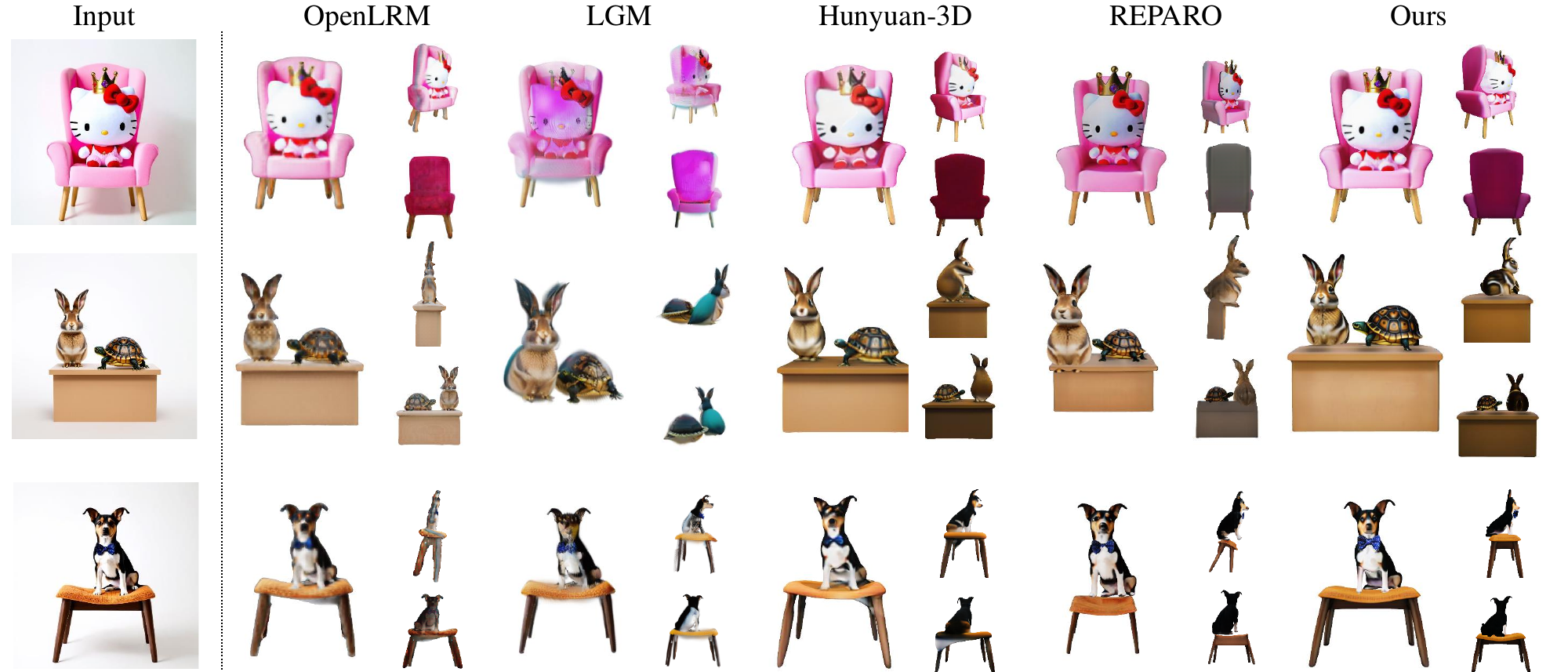}
    \vspace{-.2in}
    \caption{Visual comparison results with other image-to-3D methods on our validation set. Other methods always lead to confusion of object structure and logical mismatch, while ours can generate high visual results. We use Hunyuan-3D as our single object reconstruction model here.}
    \label{fig:compare}
    \vspace{-.1in}
\end{figure*}

\begin{figure}[t]
    \centering
    \includegraphics[width=1.0\linewidth]{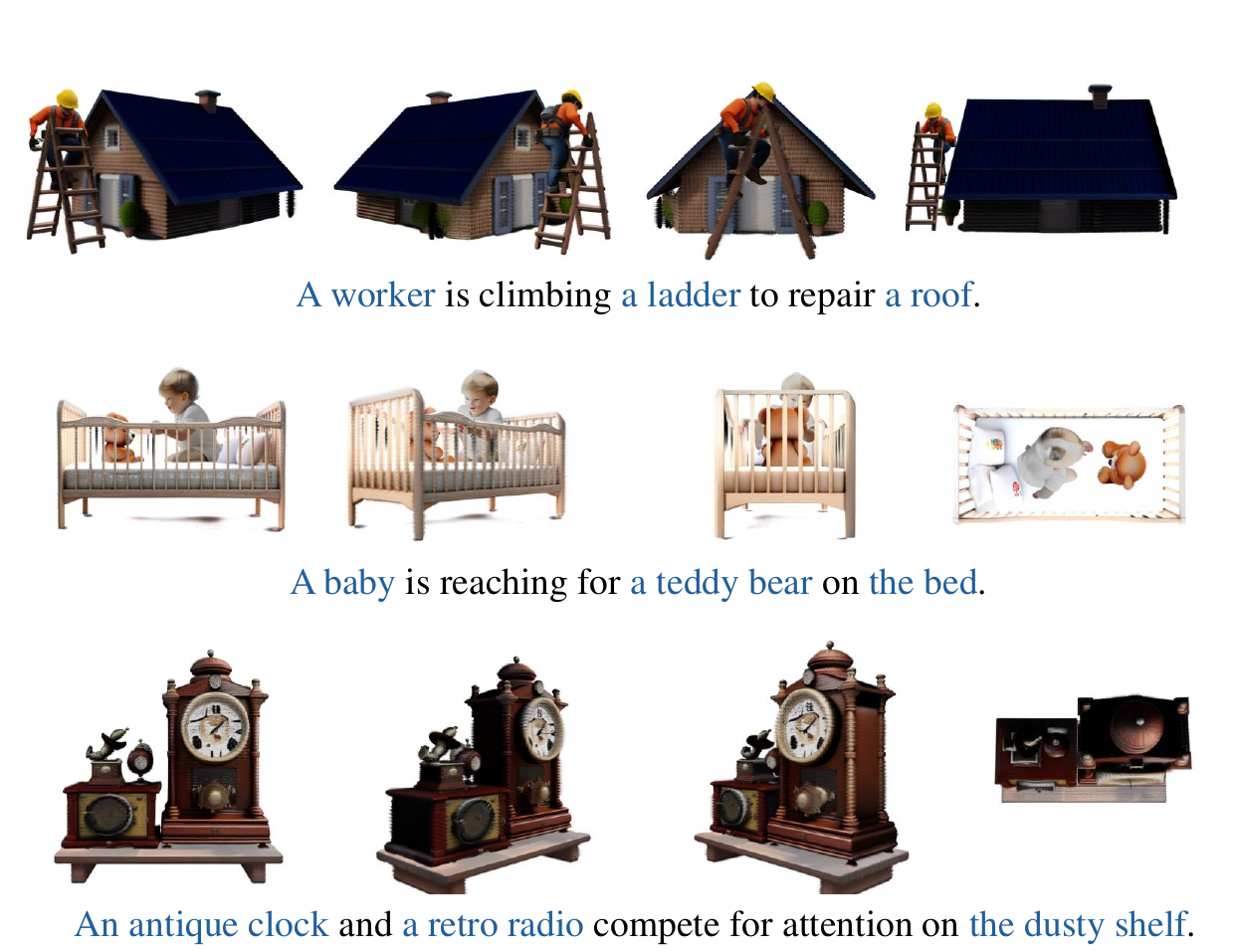}
    \vspace{-.2in}
    \caption{Our results on {T$^{3}$Bench}. Our method can generate exquisite and informative 3D assets according to the prompts.}
    \label{fig:T3show}
    \vspace{-.1in}
\end{figure}

\begin{figure*}
    \centering
    \includegraphics[width=1\textwidth]{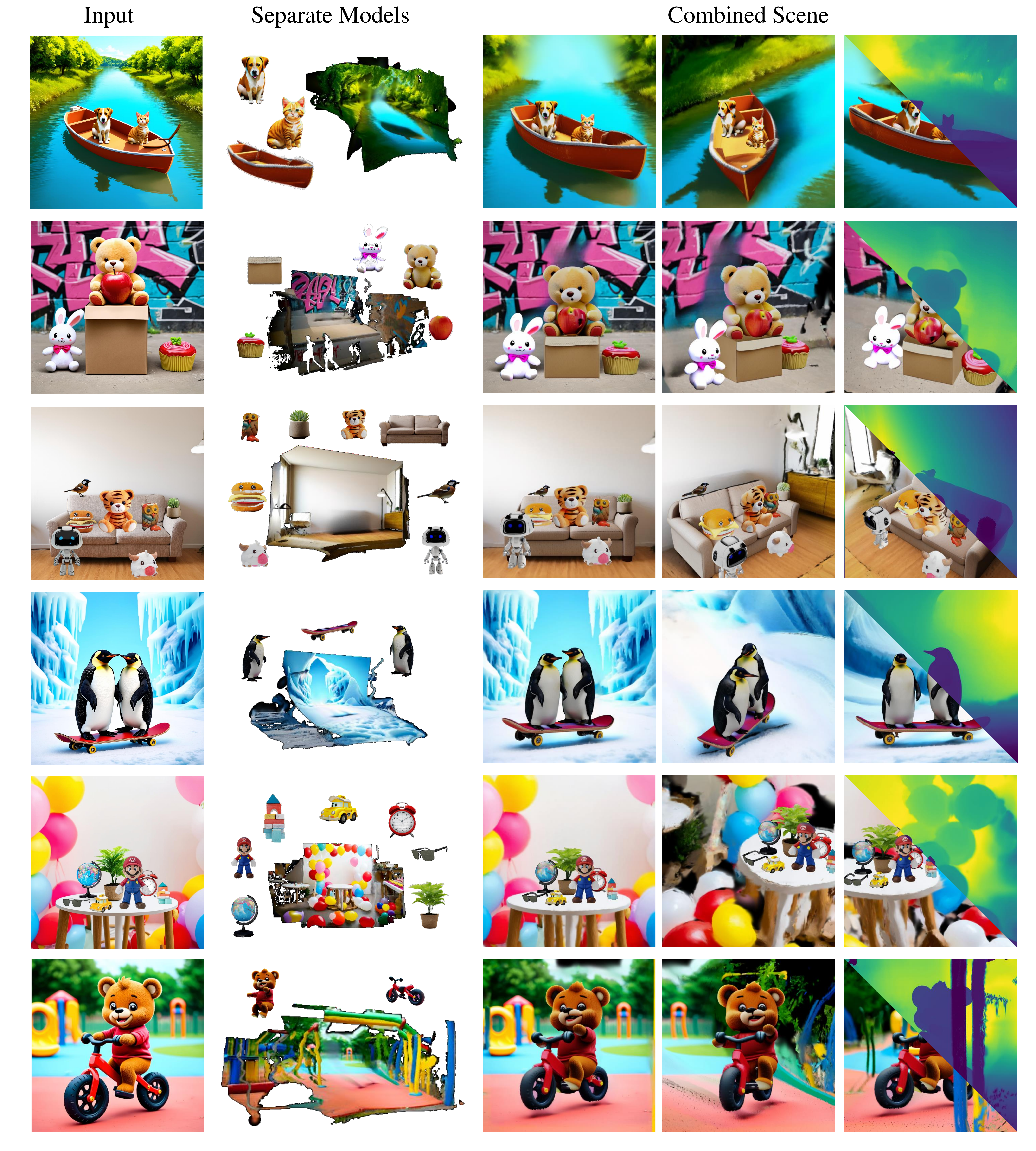}
    \vspace{-0.2cm}
    \caption{Qualitative results. \textit{Flash Sculptor} can generate high-quality 3D scene from a single image that contains multiple instances.}
    \vspace{-0.4cm}
    \label{fig:ourresults}
\end{figure*}

\subsection{Implementation Details}
We use Stable Diffusion 3 \cite{esser2024scaling} as our text-to-image generator. We set the guidance scale to 7.5 and the number of inference steps to 100 when inpainting the image. We use TRELLIS as our single-object reconstruction model in default. We set the number of elevation and azimuth \(C\) to 24 in coarse rotation stage. We set the number of points per instance \(K\) for depth alignment as 500. We set \(\lambda\) as 3 during outlier removal. Our experiments are conducted on a single NVIDIA TITAN RTX GPU and our pipeline takes approximately 4 minutes to complete a sample without a scene and 24 minutes to complete a sample with a scene.

\subsection{Main Results} 

\noindent \textbf{Benchmark.} To compare our method with other image-to-3D methods, we construct a validation set of 20 images, each containing at least two assets. The benchmark comprises 15 images generated by Stable Diffusion \cite{esser2024scaling} and 5 real-world images. We directly process these images through our pipeline to produce the final results. 
We also compare our method with other text-to-3D approaches on the multi group of {T$^{3}$Bench}\cite{he2023t}. 
In this comparison, we first utilize diffusion models \cite{esser2024scaling,li2024hunyuan} to generate 2D images, and then process them through our pipeline to obtain the final results.

\noindent \textbf{Comparison Methods.} We first compare our approach with four image-to-3D methods. Among them, OpenLRM \cite{hong2023lrm}, LGM \cite{tang2024lgm}, and Hunyuan-3D \cite{yang2024hunyuan3d} are traditional image-to-3D reconstruction methods, while REPARO \cite{han2024reparo} is a compositional 3D generation method. Additionally, we compare our method against eight text-to-3D approaches: DreamFusion \cite{poole2022dreamfusion}, SJC \cite{wang2023score}, Latent-NeRF \cite{metzer2023latent}, Magic3D \cite{lin2023magic3d}, MVDream \cite{shi2023mvdream}, and ProlificDreamer \cite{wang2023prolificdreamer} represent traditional text-to-3D techniques, whereas LayoutDreamer \cite{zhou2025layoutdreamer} and CompGS \cite{ge2024compgs} employ compositional text-to-3D generation strategies. Because all these methods only reconstruct 3D objects, we halt our pipeline at object composition, omitting scene interpolation for a fair comparison.

\noindent \textbf{Qualitative Comparison.} As shown in Figure~\ref{fig:compare}, compared with other image-to-3D methods, our method can generate compositional 3D scene with better visual quality and layout plausibility. Figure~\ref{fig:T3show} shows the performance of our approach on some of the prompts in {T$^{3}$Bench}(multi). More qualitative results of our method integrated with the scene can be seen in Figure~\ref{fig:ourresults}.

\noindent \textbf{Quantitative Comparison with Image-to-3D Approaches.} We compare the visual quality using PSNR, SSIM \cite{wang2004image} and LPIPS \cite{zhang2018unreasonable}. We also involve CLIP-Score \cite{radford2021learning} to measure semantic similarities between novel-view images and the reference image, and BLIP-VQA score \cite{huang2023t2i} to incorporate text-aligned assessment. Tab.~\ref{tab:quantitative_comparison} shows that our method outperforms others in visual similarity, multi-view consistency and semantic alignment.

\begin{table}[t]
    \centering
    \caption{Quantitative comparison with image-to-3D methods on our benchmark.}
    \renewcommand{\arraystretch}{1.2} 
    \resizebox{\columnwidth}{!}{ 
        \begin{tabular}{l @{\hspace{10mm}}c @{\hspace{6mm}}c @{\hspace{6mm}}c @{\hspace{6mm}}c @{\hspace{6mm}}c}
            \toprule[1pt]
            \textbf{Method} & \textbf{CLIP - Score$\uparrow$} & \textbf{BLIP - VQA Score$\uparrow$} & \textbf{PSNR$\uparrow$} & \textbf{SSIM$\uparrow$} & \textbf{LPIPS$\downarrow$} \\
            \midrule
            OpenLRM \cite{hong2023lrm} & 81.93\% & 26.17\% & 13.055 & 0.773 & 0.303 \\
            LGM \cite{tang2024lgm} & 84.33\% & 34.98\% & 12.420 & 0.761 & 0.328 \\
            Hunyuan-3D \cite{yang2024hunyuan3d} & 86.90\% & 40.52\% & 13.136 & 0.791 & 0.301 \\
            REPARO \cite{han2024reparo} & 87.61\% & 38.18\% & 12.850 & 0.781 & 0.312 \\
            Ours & \textbf{90.31}\% & \textbf{46.83\%} & \textbf{15.310} & \textbf{0.836} & \textbf{0.275} \\
            \bottomrule[1pt]
        \end{tabular}
    }
    \vspace{-0.1cm}
    \label{tab:quantitative_comparison}
\end{table}

\noindent \textbf{Quantitative Comparison with Text-to-3D Approaches.}  
We evaluate our method using the two metrics provided by {T$^{3}$Bench}: quality and alignment. As shown in Tab.~\ref{tab:t3bench}, our approach consistently surpasses existing methods across both metrics while maintaining a much lower computational cost.

\begin{table}[t]
    \centering
    \caption{Quantitative comparison on \text{T$^{3}$Bench} with other text-to-3D methods. Our method requires 4 minutes on a single NVIDIA TITAN RTX and 2 minutes on a single H800 respectively. }
    \renewcommand{\arraystretch}{1.2} 
    \resizebox{\linewidth}{!}{ 
    \begin{tabular}{l @{\hspace{10mm}}c@{\hspace{6mm}} c@{\hspace{6mm}} c@{\hspace{6mm}} c}
        \toprule[1pt]
        \textbf{Method} & \textbf{Time} & \multicolumn{3}{c}{\textbf{T$^{3}$Bench (Multiple Objects)}}  \\
        & & \textbf{Quality$\uparrow$} & \textbf{Alignment$\uparrow$} & \textbf{Average$\uparrow$} \\
        \midrule
        DreamFusion \cite{poole2022dreamfusion} & 30 minutes & 17.3 & 14.8 & 16.1 \\
        SJC \cite{esser2024scaling} & 25 minutes & 17.7 & 5.8 & 11.7 \\
        Latent-NeRF \cite{metzer2023latent} & 65 minutes & 21.7 & 19.5 & 20.6 \\
        Magic3D \cite{lin2023magic3d} & 40 minutes & 26.6 & 24.8 & 25.7 \\
        MVDream \cite{shi2023mvdream} & 30 minutes & 39.0 & 28.5 & 33.8 \\
        ProlificDreamer \cite{wang2023prolificdreamer} & 240 minutes & 45.7 & 25.8 & 35.8 \\
        \midrule
        LayoutDreamer \cite{zhou2025layoutdreamer} & 60 minutes & 56.6 & 31.8 & 44.2 \\
        CompGS \cite{ge2024compgs} & 70 minutes & 54.2 & 37.9 & 46.1 \\
        \midrule
        \multirow{2}{*}{\textbf{{\scshape Ours}}} & 4 minutes (Titan RTX) & \multirow{2}{*}{{(\textcolor{red}{+0.2})} \textbf{56.8}} & \multirow{2}{*}{{(\textcolor{red}{+0.3})} \textbf{38.2}} & \multirow{2}{*}{{(\textcolor{red}{+1.4})} \textbf{47.5}} \\
            & 2 minutes (H800) &  &  &  \\                            
        \bottomrule[1pt]
    \end{tabular}
    }
    \label{tab:t3bench}
\end{table}

\begin{table}[t]
    \centering
    \caption{Quantitative comparison with image-to-3D methods on our benchmark.}
    \renewcommand{\arraystretch}{1.2} 
    \resizebox{\columnwidth}{!}{ 
        \begin{tabular}{l @{\hspace{10mm}}c @{\hspace{6mm}}c @{\hspace{6mm}}c @{\hspace{6mm}}c @{\hspace{6mm}}c}
            \toprule[1pt]
            \textbf{Method} & \textbf{CLIP - Score$\uparrow$} & \textbf{BLIP - VQA Score$\uparrow$} & \textbf{PSNR$\uparrow$} & \textbf{SSIM$\uparrow$} & \textbf{LPIPS$\downarrow$} \\
            \hline
           Full  &  \textbf{90.31}\% & \textbf{46.83\%} & \textbf{15.310} & \textbf{0.836} & \textbf{0.275}\\
    w/o rotation refinement    & 90.22\% & 46.80\% & 15.184 & 0.828 & 0.279\\
    w/o outlier removal    & 89.32\% & 45.98\% & 14.856 & 0.825 & 0.288\\
        \bottomrule[1pt]
        \end{tabular}
    }
    \label{tab:ablation}
\end{table}

\subsection{Ablation Studies}

\begin{figure}[t]
    \centering
    \includegraphics[width=1.0\linewidth]{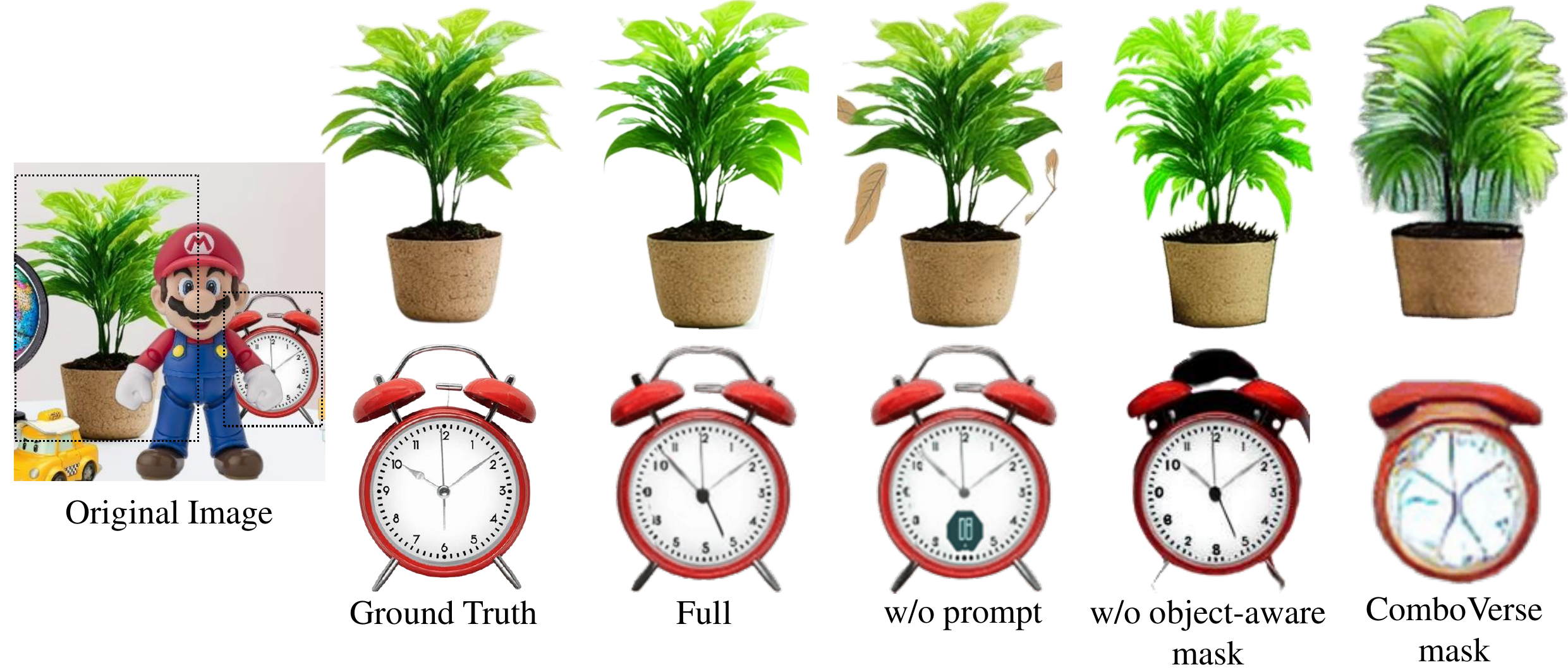}
    \vspace{-.3in}
    \caption{Ablation experiments on mask generation for object inpainting.}
    \label{fig:ablation1}
    \vspace{-.1in}
\end{figure}

\noindent \textbf{Ablation Studies for mask generation for object inpainting.} We conducted ablation studies to evaluate the effectiveness of mask for object inpainting. We first removed our prompt and negative prompt. As in Fig. \ref{fig:ablation1}, some unrelated parts were generated. Then we set the inpainting mask as all the part excluding the object mask within the bounding box, we can see that more extra parts were generated. We also compared our inpainting mask with that used in ComboVerse \cite{chen2024comboverse}, and observed that the object structures were significantly altered in this case.

\begin{figure}[t]
    \centering
    \includegraphics[width=1.0\linewidth]{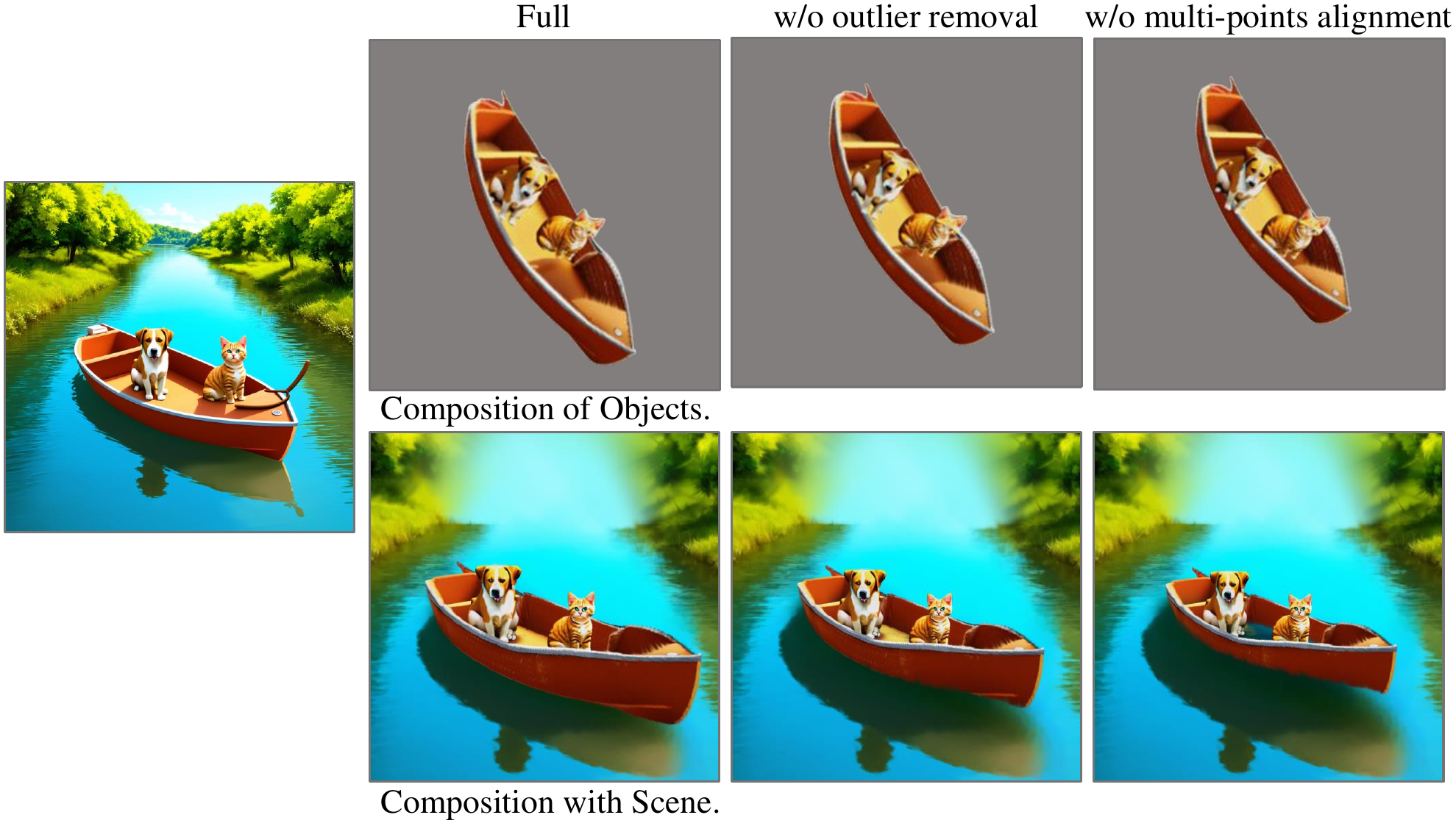}
    \vspace{-.2in}
    \caption{Ablation experiments on depth piror.}
    \label{fig:ablation2}
    \vspace{-0.05in}
\end{figure}

\noindent \textbf{Ablation Studies for depth piror.} To evaluate the effectiveness of our strategies on depth translation, i.e., outlier removal and multi-point alignment, we conducted experiments on both objects composition and scene composition. As shown in Fig. \ref{fig:ablation2}, outlier removal can further accurately adjust the depth position and only using the mid point to align will degrade the performance significantly.

\begin{figure}
    \centering
    \includegraphics[width=1.0\linewidth]{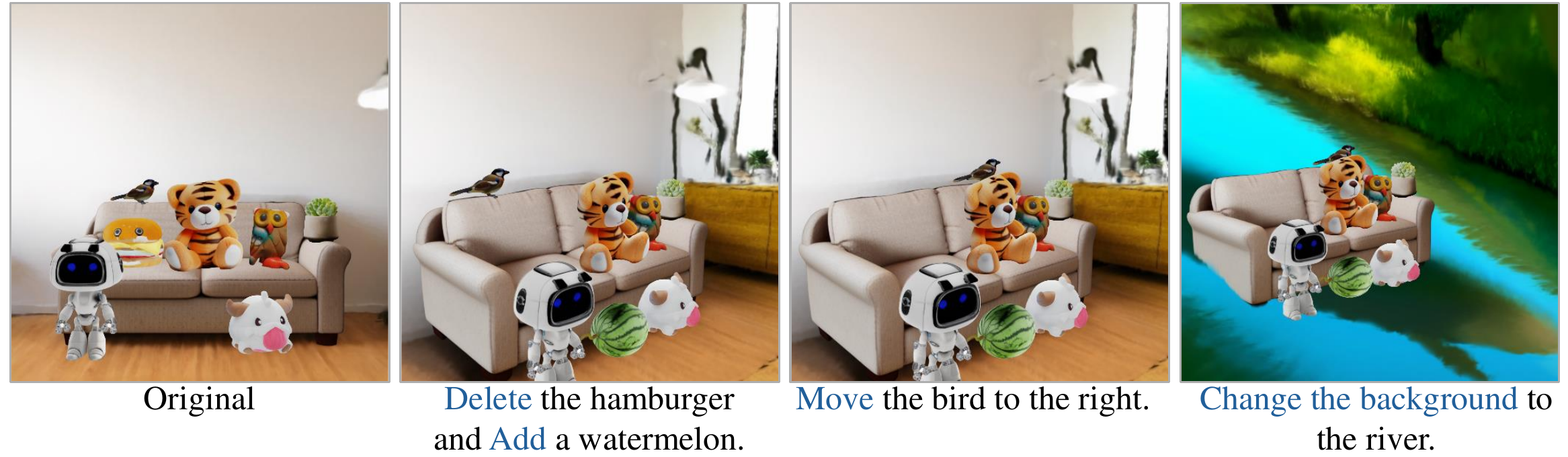}
    \vspace{-.2in}
    \caption{An example of 3D editing application.}
    \label{fig:ed}
    \vspace{-.2in}
\end{figure}

\noindent \textbf{Quantitative Ablation Studies.} We conducted quantitative ablation studies on the rotation strategy and depth alignment strategy of our method on our validation dataset. We can see from Tab.~\ref{tab:ablation} that the performance of the pipeline will be degraded without rotation refinement and outlier removal.

\subsection{Applications} 
\noindent \textbf{3D Editing.} It is straightforward and flexible for our method to perform various 3D editing operations. As shown in Fig. \ref{fig:ed}, these editing operations enable intuitive and efficient manipulation of complex 3D scenes.

\section{Conclusions}

In this paper, we introduce Flash Sculptor, a novel framework for compositional 3D scene and object reconstruction from a single image. 
By decomposing the complex task of 3D reconstruction into independent sub-tasks, our approach circumvents the cumbersome global layout optimization process required by existing compositional 3D methods.
Through experiments, we can see that Flash Sculptor significantly improves both accuracy and computational efficiency compared to state-of-the-art text-to-3D and image-to-3D models.
These results highlight the promise of our divide-and-conquer strategy for scalable and practical compositional 3D scene synthesis.

    \small
    \bibliographystyle{ieeenat_fullname}
    \bibliography{main}

\end{document}